\documentclass[10pt,twocolumn]{article}

\usepackage{iccv}
\usepackage{times}
\usepackage{epsfig}
\usepackage{graphicx}
\usepackage{amsmath}
\usepackage{amssymb}
\usepackage{subcaption}
\usepackage{booktabs}

\usepackage{fancybox}
\usepackage{xcolor}

\usepackage[pagebackref=true,breaklinks=true,bookmarks=false]{hyperref}

\iccvfinalcopy 

\def\iccvPaperID{19} 

\ificcvfinal\fi
\begin{document}

\title{Count-ception: Counting by Fully Convolutional Redundant Counting}

\author{Joseph Paul Cohen\\
Montreal Institute for Learning Algorithms\\
Universit\'{e} of Montr\'{e}al\\
Friends of the Farlow Fellow\\
Harvard University Herbaria \\
{\tt\small cohenjos@iro.umontreal.ca}
\and
Genevi\`{e}ve Boucher\\
Institute for Research in Immunology and Cancer\\
Universit\'{e} of Montr\'{e}al\\
{\tt\small genevieve.boucher@umontreal.ca}
\and
Craig A. Glastonbury\\
Big Data Institute\\
University of Oxford\\
{\tt\small craig@well.ox.ac.uk}
\and
Henry Z. Lo\\
Department of Computer Science\\
University of Massachusetts Boston\\
{\tt\small henryzlo@cs.umb.edu}
\and
Yoshua Bengio\\
CIFAR Senior Fellow\\
Montreal Institute for Learning Algorithms\\
Universit\'{e} of Montr\'{e}al\\
{\tt\small yoshua.bengio@umontreal.ca}
}

\maketitle

\begin{abstract}

Counting objects in digital images is a process that should be replaced by machines. This tedious task is time consuming and prone to errors due to fatigue of human annotators. The goal is to have a system that takes as input an image and returns a count of the objects inside and justification for the prediction in the form of object localization. We repose a problem, originally posed by Lempitsky and Zisserman, to instead predict a count map which contains redundant counts based on the receptive field of a smaller regression network. The regression network predicts a count of the objects that exist inside this frame. By processing the image in a fully convolutional way each pixel is going to be accounted for some number of times, the number of windows which include it, which is the size of each window, (i.e., 32x32 = 1024). To recover the true count we take the average over the redundant predictions. Our contribution is redundant counting instead of predicting a density map in order to average over errors. We also propose a novel deep neural network architecture adapted from the Inception family of networks called the Count-ception network. Together our approach results in a 20\% relative improvement (2.9 to 2.3 MAE) over the state of the art method by Xie, Noble, and Zisserman in 2016.

\end{abstract}

\begin{figure}
\begin{center}
  \includegraphics[width=0.95\columnwidth]{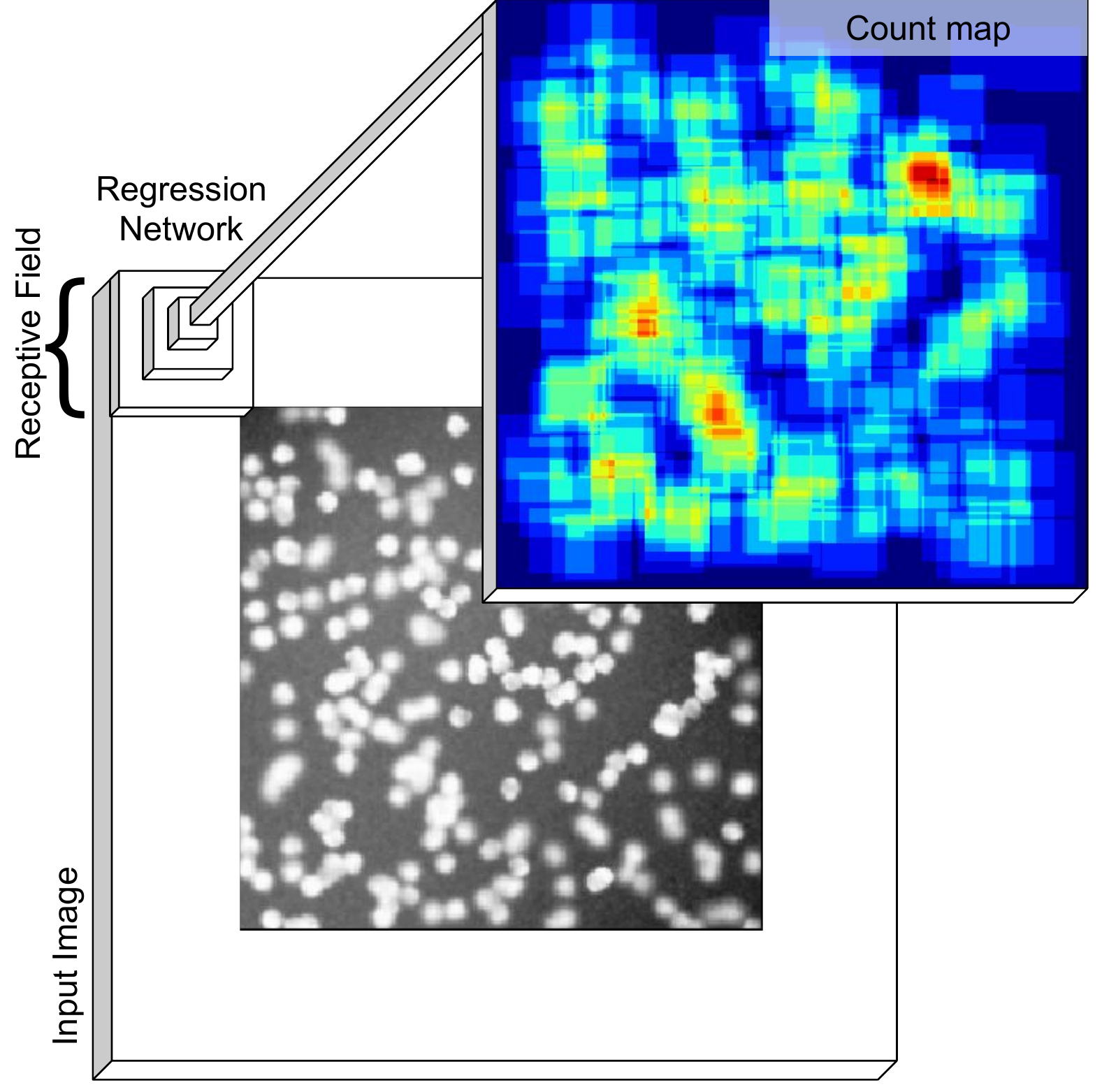}
  \caption{Given an image, the regression network counts the number of objects in each receptive field. The predicted count map corresponds to the receptive field of the regression network. The upper left pixel of the activation map is based on only one pixel of the input image in the upper left corner.}
  \label{fig:network1}
  \vspace{-10pt}
\end{center}
\end{figure}

\section{Introduction}

Counting objects in digital images is a process that is time consuming and prone to errors due to fatigue of human annotators. The goal of this research area is to have a system that takes as input an image and returns a count of the objects inside and justification for the prediction in the form of object localization.

The classical approach to counting involves fine-tuning edge detectors to segment objects from the background \cite{Sankur2004} and counting each one.  A large challenge here is dealing with overlapping objects which require methods such as the watershed transformation \cite{Beucher1994}. These approaches have many hyperparameters specifically for each task and are complicated to build. 

The core of modern approaches was described by Lempitsky and Zisserman in 2010 \cite{Lempitsky2010}. Given labels with point annotations of each object, they construct a density map of the image. Here, each object predicted takes up a density of 1, so a sum of the density map will reveal the total number of objects in the image. This method naturally accounts for overlapping objects We extend this idea and focus on two main areas:

\begin{enumerate}
\item We propose \textit{redundant counting} instead of a density map approach in order to average over errors.
\item We propose a novel construction of networks and training that can apply to counting tasks with very complicated objects.
\end{enumerate}

\begin{figure}
\begin{center}
  \includegraphics[width=0.95\columnwidth]{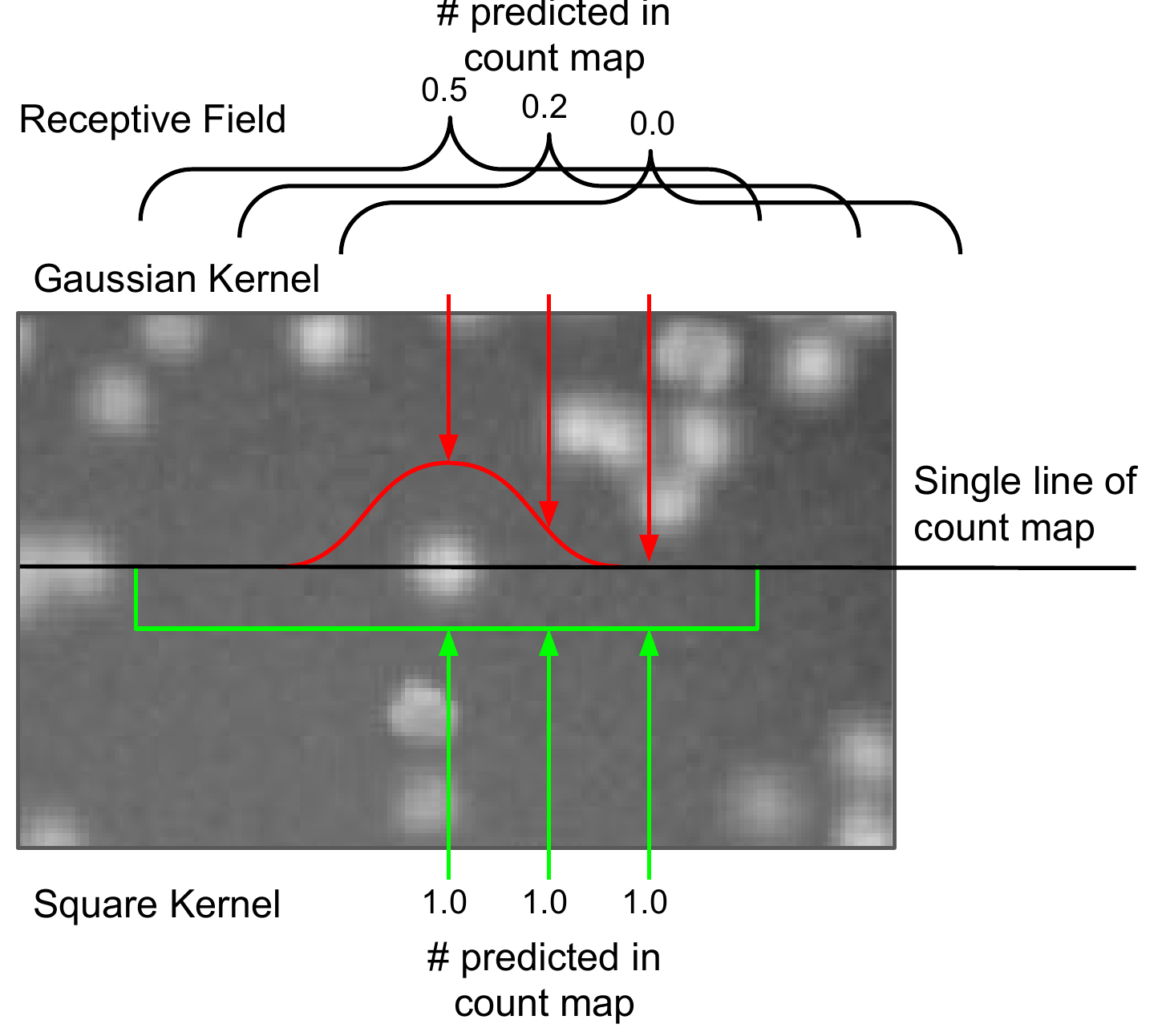}
  \caption{Comparing how a single row of the count map can be calculated for single cell. Above the line in red are the values that the network is trained to predict when a Gaussian kernel is used. Below in green are the values when the square kernel is used. The square kernel is the same size as the receptive field.}
  \label{fig:kernel}
  \vspace{-20pt}
\end{center}
\end{figure}

\begin{figure}[th]
\begin{center}

\includegraphics[width=0.5\columnwidth]{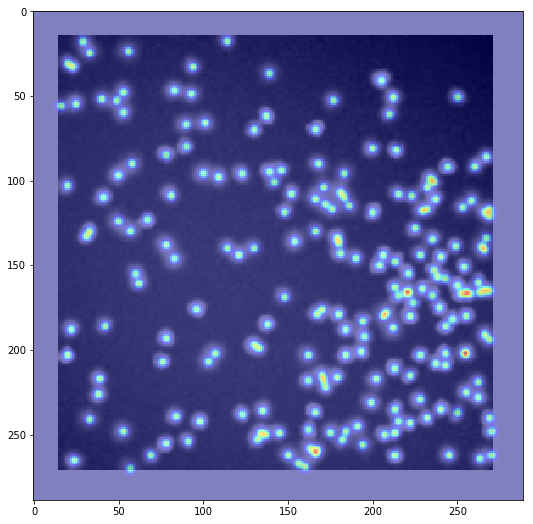}%
\includegraphics[width=0.5\columnwidth]{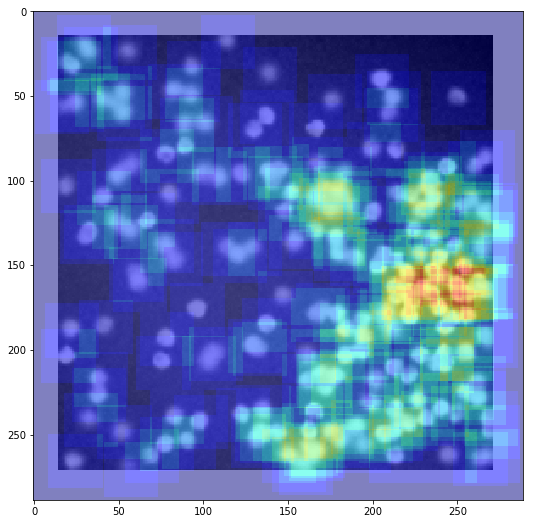}%

  \caption{Comparison between annotations using Gaussian and square kernels. }
  \label{fig:gausvsq}
\end{center}
\end{figure}

We repose the problem of predicting a density map to instead predict a count map which contains \textit{redundant counts} based on the receptive field of a smaller regression network. The regression network predicts a count of the objects that exist inside this frame as shown in Figure \ref{fig:network1}.
By processing the image in a fully convolutional way \cite{Long2015} each pixel is going to be accounted for some number of times, the number of windows which include it, which is the size of each window, (i.e., $32 \times 32 = 1024$). To recover the true count we can take the average of all these predictions. Figure \ref{fig:kernel} illustrates how this change in kernel makes more sense with respect to the receptive field of the network that must make predictions. Using the Gaussian density map forces the model to predict specific values based on how far the cell is from the center of the receptive field. This is a harder task than just predicting the existence of the cell in the receptive field.  A comparison of these two types of count maps is shown in Figure \ref{fig:gausvsq}.

To perform this prediction we focus on a method using deep learning \cite{LeCun2015} and convolutional neural networks \cite{LeCun1998} like Xie \cite{Xie2016} and Arteta \cite{Arteta16} have. They utilized networks similar to FCN-8 \cite{Long2015} which form bottlenecks at the core of the network to capture complex relationships in different parts of the image. Instead, we pad the borders of the input image so that the receptive field of the regression network will redundantly count the correct number of times. This way we do not bottleneck the representation in any way. 

\section{Related Work}
  
The idea of counting with a density map began with Lempitsky and Zisserman in 2010 \cite{Lempitsky2010} where they used dense SIFT features from the image as input to a linear regression to predict a density map. We predict redundant counts instead of a density map. Although a summation over the output of the model is taken over both causes, our method is explicitly designed to tolerate the errors when predictions are made.

However, the density map of objects does count multiple times indirectly. It needs to properly predict a density map of objects which is generated from a small Gaussian with the mean at the point annotation. The values they need to predict vary as some are at the mean and some are not. It doesn't take into account the receptive field so the objects may be in view and the network has to suppress its prediction.

Many approaches were introduced to predict a better density map. Fiaschi 2012 \cite{Fiaschi2012} used a regression forest instead of a linear model to make the density prediction based on BoW-SIFT features. Arteta 2014 \cite{Arteta2014} proposed an interactive counting algorithm which would extend this algorithm to more dynamically learn to count various concepts in the image. Xie 2016 \cite{Xie2016} introduced deep neural networks to this problem. Their method built a network which would convolve a $100 \times 100$ region to a $100 \times 100$ density map. Once this network was trained it can be run in a fully convolutional way similar to our method. However, these approaches focus on predicting a density map which differentiates them from our work.

Arteta 2016 \cite{Arteta16} discuss new approaches past the density model. Their focus is different than our work. They tackle the problem of incorporating multiple point annotations from noisy crowd sourced data. They also utilize segmentation of the background to filter our erroneous predictions that may happen there.

In Segui \cite{Segui2015} their method takes the entire image as input and output a single count value using fully connected layers to break the spatial relationship. They discover that a network can learn to count and while doing this they learn features for identifying the objects such as MNIST digits. We use this idea in that the regression network is learning to count the $32 \times 32$ frame. But we expect it to produce errors so we perform this task redundantly.

Xie in 2015 \cite{Xie2015} presented an interesting idea similar to the direction we are going in. Their goal is to predict a proximity map which consists of cone shaped distributions over each cell which smooths each cell prediction using surrounding detections. This cone extended only 5 pixels from the point annotation which was the average size of the cell. However, this approach is more in line with a density map than a count map.

\section{Fully Convolutional Redundant Counting}

\begin{table}[]
\begin{center}
\caption{Notation used in this paper.}
\label{tab:notation}
\begin{tabular}{l l}
\toprule
Symbol & Description \\
\midrule
$I$ & input image \\
$T$ & target image, constructed from $L$ \\
$L$ & image of point notations \\
$s$ & stride length \\
$r$ & width / length of receptive field \\
$R(x,y)$ & receptive field associated with $x,y$ \\
$F(I)$ & map of predicted counts for $I$ \\
$N$ & number of training / validation images \\
\bottomrule
\end{tabular}

\end{center}
\end{table}

\subsection{Problem Statement}

We would like to obtain the count of objects in an input image $I$ being given only a few training examples with point annotations of each object. The objects to count are often very small, and the overall image very large.  Because counting is labor-intensive, there are often few labeled images in practice.


\begin{figure}
\begin{center}

  \includegraphics[width=1.0\columnwidth]{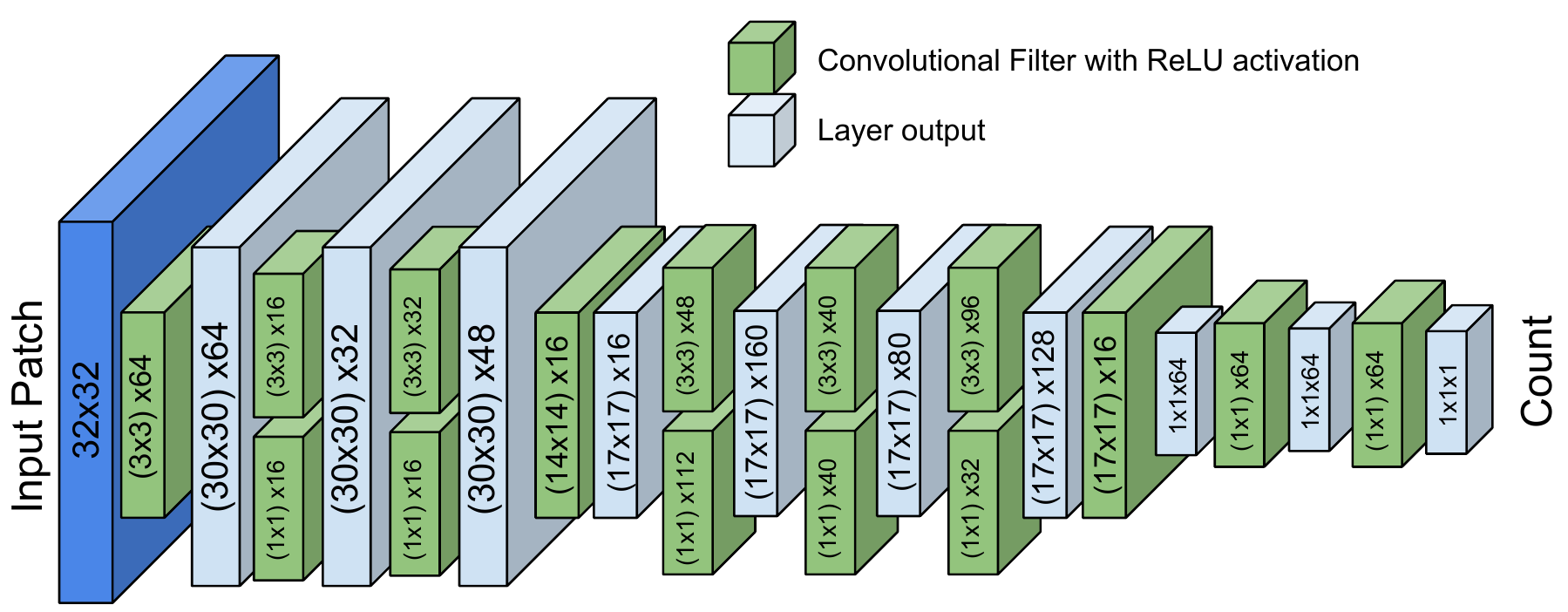}
  \caption{The Count-ception network architecture that is used for the regression network. Each intermediate tensor is labeled (filter size) x \# filters There are two points in the network where the size is reduced. The $3 \times 3$ convolutions are padded so they do not reduce the size. Batch Normalization layers are inserted after each convolution but not pictured here.}
  \label{fig:network-arch}
\end{center}
\end{figure}

\subsection{Overview of Technique}

\textbf{Motivation}: We want to merge the idea of networks that count everything in their receptive field by Segui \cite{Segui2015} with the density map of objects by Lempitsky and Zisserman \cite{Lempitsky2010} using fully convolutional processing like Xie \cite{Xie2016} and Arteta \cite{Arteta16}.

\textbf{Technique}: Instead of using a CNN that takes the entire image as input and produces a single prediction for the number of objects we use a smaller network that is run over the image to produce an intermediate count map. This smaller network is trained to count the number of objects in its receptive field. More formally; we process the image $I$ with this network in a fully convolutional way to produce a matrix $F(I)$ that represents the counts of objects for a specific receptive field $r \times r$ of a sub-network that performs the counting. A high-level overview:

\begin{enumerate}
\item Pre-process image by padding
\item Process image in a fully convolutional way
\item Combine all counts together into total count for image
\end{enumerate}

The fully convolutional network processes an image by applying a network with a small receptive field on the entire image.  This has two effects which reduce overfitting.  First, by being small, the fully convolutional network has much fewer parameters than a network trained on the entire image.  Second, by splitting up an image, the fully convolutional network has much more training data to fit parameters on. 

\begin{figure}
\begin{center}

\includegraphics[width=1.0\columnwidth]{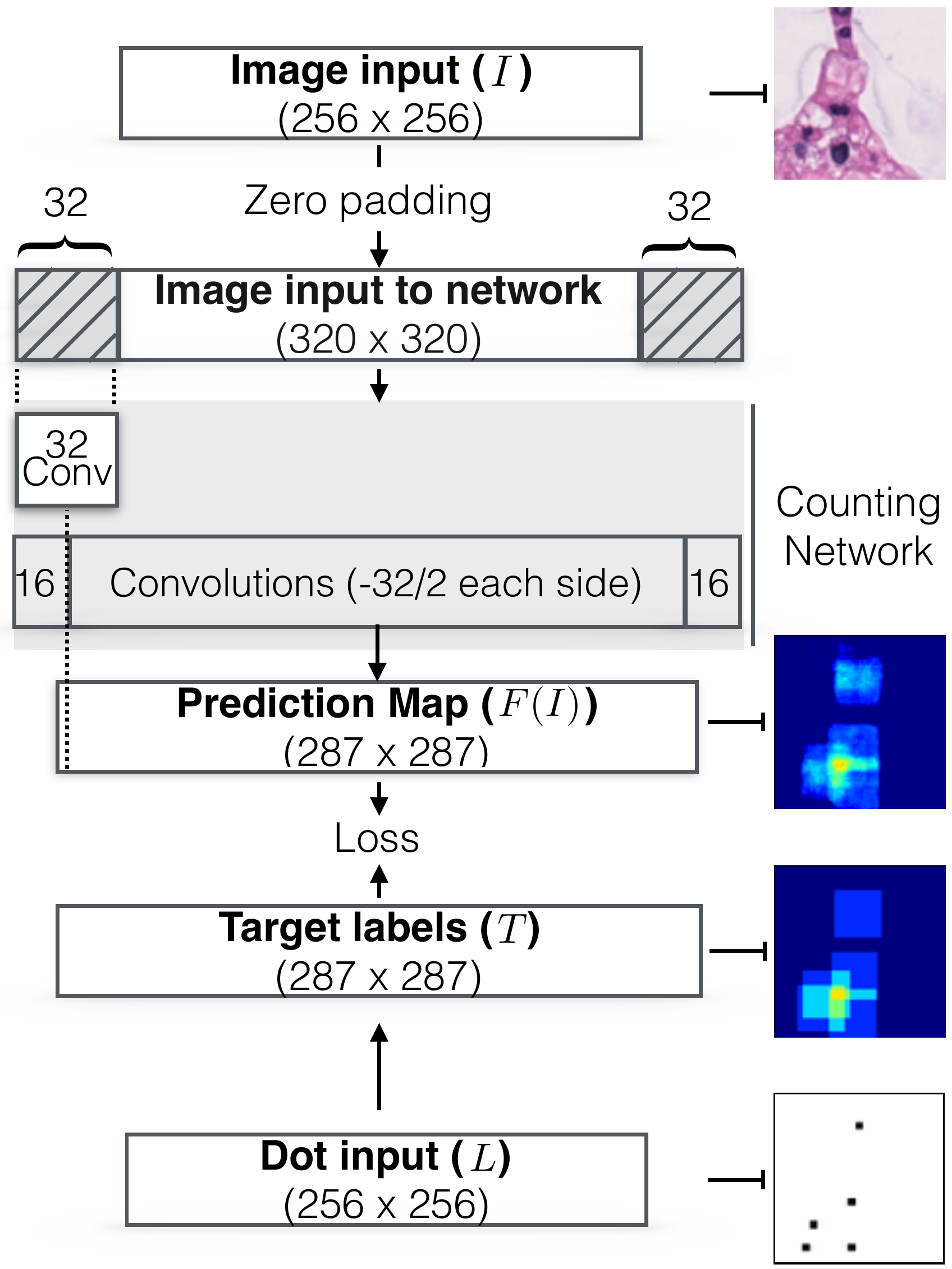}%
  \caption{Here is the pipeline for $r=32$ given an input image that is $256 \times 256$. The input image is padded and convolved to calculate the prediction count map which should match the target count map. The count map will be non-zero after $r/2$ from the border of the input image. A loss is calculated between the prediction count map and target count map in order to update the weights of the counting network to better match the target count map.}
  \label{fig:shapes}
\end{center}
\end{figure}

The following discussions will consider a receptive field of $32$ for simplicity and in order to have concrete examples. This method can be used with any receptive field size. An overview of the process is shown in Figure \ref{fig:shapes}.

\subsection{Input}

We want to count target objects in an image $I$.  This image has multiple target objects that are labelled with single point labels $L$.

Because the counting network only reduces the dimensions from $(32 \times 32) \rightarrow  (1 \times 1)$ the input $I$ must be padded in order to deal with objects that appear on the border. Objects on the border of the image will at most be in the receptive field of a network with only one column or row overlapping the input image. For $r=32$ a pixel in $F(I)$ can only be 15 pixels from the border of $I$. 

$F(I)$ is meant to align with the target $T$. It is important that these be aligned such that the receptive field of the network aligns with the proper regression target.

\subsubsection{Constructing the target image $T$}

The target image can be constructed from a point-annotated map $L$, the same size as the input image $I$, where each object is annotated by a single pixel. This is desirable because labeling with dots is much easier than drawing the boundaries for segmentation.



Let $R(x,y)$ be the set of pixel locations in the receptive field corresponding to $T[x,y]$.  Then we can construct the target image $T$:

\begin{equation}
T[x,y]=\sum_{(x',y') \in R(x,y)} L[x',y']
\end{equation}

Here $T[x,y]$ is the sum of cells contained in a region the size of the $r \times r$ receptive field. This will become the regression target for the $r \times r$ region of the image.





\subsection{Fully Convolutional Redundant Counting}


We use fully convolutional networks with a receptive field of $32 \times 32$.  The output of the fully convolutional network on the entire $320 \times 320$ image is $287 \times 287$ pixels.  This yields a fully convolutional network output image larger than the original input.  Each pixel in the output will represent the count of targets in that receptive field.

To perform this mapping we propose the Count-ception architecture which is adapted from the Inception family of networks by Szegedy et. al. \cite{Szegedy2015a}. Our proposed model is shown in Figure \ref{fig:network-arch}. At the core of the model Inception units are used to perform 1x1 (pad 0) and 3x3 (pad 1) convolutions at multiple layers without reducing the size of the tensor. After every convolution a Leaky ReLU activation is applied \cite{Maas}. We notice an improvement of the regression predictions with the Leaky ReLU during training because the output can be pushed to zero and then recover to predict the correct count.

Our modifications are in the down sampling layers. We removed the max pooling and stride=2 convolutions. They are replaced by large convolutions. This makes it easier  to calculate the receptive field of the network because strides add a modulus to the calculation of the count map size.

We perform this down sampling in two locations using large filters to greatly reduce the size of the tensor. A necessity in allowing the model to train is utilizing Batch Normalization layers \cite{Ioffe2015} after every convolution. 

\subsection{Loss Functions and Regularization}

We tried many combinations of loss functions and found $L1$ loss to perform the best. 
\begin{equation}
\min||F(I) - T||_1
\end{equation}

Xie found that the $L2$ penalty was too harsh to the network during training.  We reached the same conclusion for our configuration and chose an $L1$ loss instead. We also tried to combine this basic pixel-wise loss with a loss based on the overall prediction in the entire image. We found this caused over-fitting and provided no assistance in training. The network would simply learn artifacts in each image in order to correctly predict the overall counts. 

\subsection{Combining Sub-Image Counts}

The above loss is a surrogate objective to the real count that we want. We intentionally count each cell multiple times in order to average over possible errors.  With a stride of 1, each target is counted once for each pixel in its receptive field. As the stride increases, the number of redundant counts decreases.%
\begin{equation}
\text{\# redundant counts} = \left(\frac{r}{s}\right)^2
\end{equation}

In order to recover the true count we divide the sum of all pixels by the number of redundant counts.

\begin{equation}
\text{\# true counts} = 
\frac{\sum_{x,y} F(I)[x,y]}{
\text{\# redundant counts}}
\end{equation}

There are many benefits to using redundant counts. If the pixel label is not exactly at the center of the cell, or even outside the cell, the network can still learn because on average the cell will appear in the receptive field.

\subsection{Limitations}

\begin{figure*}
    \centering
    \begin{subfigure}[b]{1.0\textwidth}
        \includegraphics[width=\columnwidth]{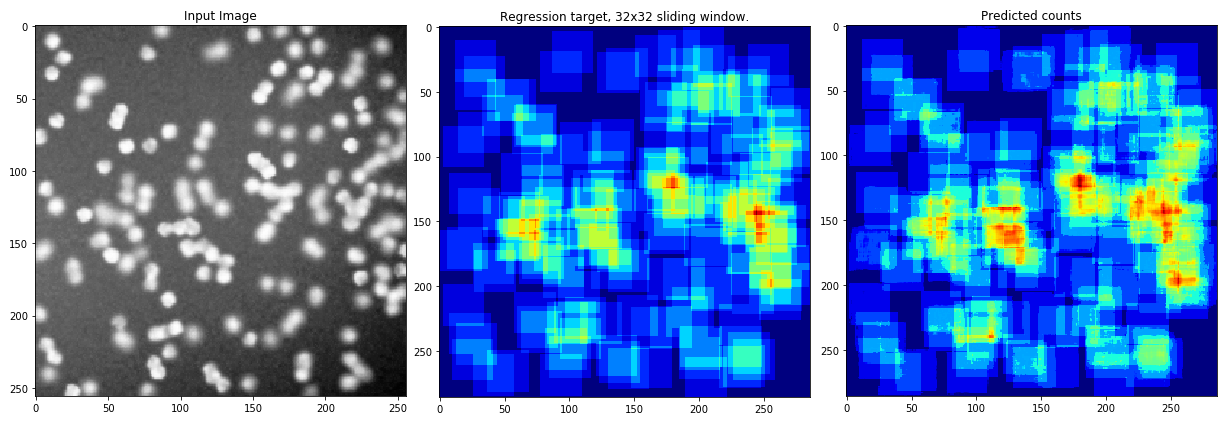}
        \vspace{-15pt}
        \caption{VGG Cell detection}
    \end{subfigure}
    
    \vspace{12pt}
    \begin{subfigure}[b]{1.0\textwidth}
        \includegraphics[width=\columnwidth]{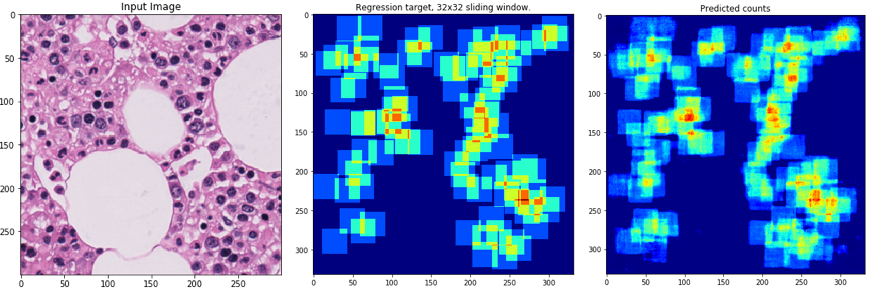}
        \vspace{-15pt}
        \caption{MBM Cell detection}
    \end{subfigure}
    
    \vspace{12pt}
    \begin{subfigure}[b]{1.0\textwidth}
        \includegraphics[width=\columnwidth]{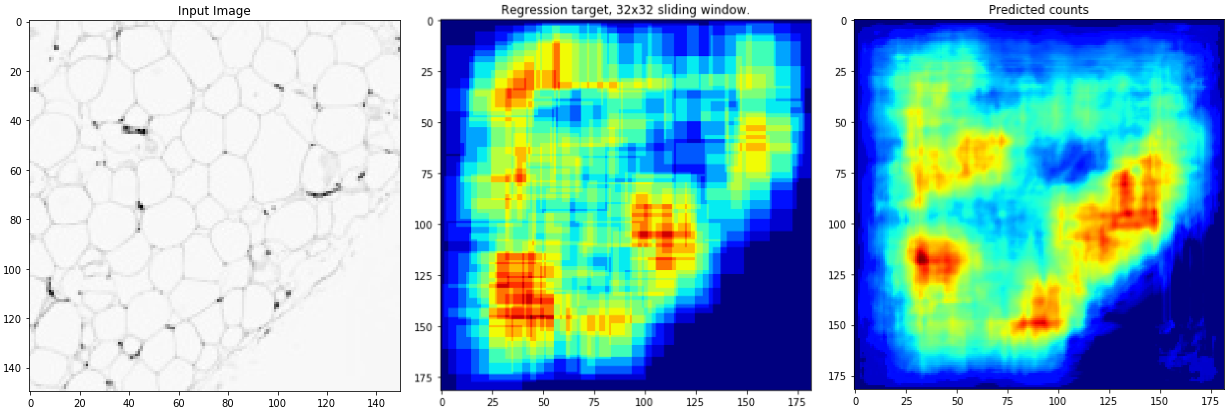}
        \vspace{-15pt}
        \caption{Adipocyte Cell detection}
    \end{subfigure}
    \caption{Example predicted count maps of images held out of training for each dataset. On the left is the image. In the center is the correct output which would result in the correct count of cells. On the right is the predicted count map.}
    \label{fig:detections}
\end{figure*}

With this approach we sacrifice the ability to localize each cell exactly with $x,y$ coordinates. Viewing the predicted count map can localize where the detection came from (shown in Figure \ref{fig:detections}) but not to a specific coordinate. For many applications accurate counting is more important than exact localization. Another issue with this approach is that a correct overall count may not come from correctly identifying cells and could be the network adapting to the average prediction for each regression. One common example is if the training data contains many images without cells the network may predict 0 in order to minimize the loss. A solution similar to Curriculum Learning \cite{Bengio2009} is to first train on a more balanced set of examples and then take well performing networks and train them on more sparse datasets.

\section{Datasets}

\begin{figure}[!ht]
    \centering
    \begin{subfigure}[b]{0.3\columnwidth}
        \includegraphics[width=\columnwidth]{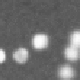}
        \caption{VGG Cells}
    \end{subfigure}
    ~
    \begin{subfigure}[b]{0.3\columnwidth}
        \includegraphics[width=\columnwidth]{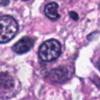}
        \caption{MBM Cells}
    \end{subfigure}
    ~
    \begin{subfigure}[b]{0.3\columnwidth}
        \includegraphics[width=\columnwidth]{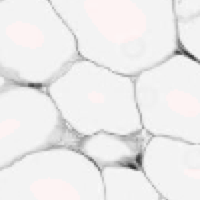}
        \caption{Adipocyte Cells}
        \label{fig:adi}
    \end{subfigure}
    \caption{Examples of cells in each dataset used for evaluation.}
\end{figure}

\textbf{VGG Cells}: To compare with the state of the art we first use the standard benchmark dataset which was introduced by Lempitsky and Zisserman in 2010 \cite{Lempitsky2010}. There are 200 images with a 256x256 resolution that contain simulated bacterial cells from fluorescence-light microscopy created by \cite{Lehmussola2007}. Each image contains 174 $\pm$ 64 cells which overlap and are at various focal distances simulating real life imaging with a microscope. 

\textbf{MBM Cells}: We also use a real dataset based on the BM dataset introduced by Kainz et al. in 2015 \cite{Kainz2015} which consists of eleven $1,200 \times 1,200$ resolution images of bone marrow from height healthy individuals. The standard staining procedure used depicts in blue the nuclei of the various cell types present whereas the other cell constituents appear in various shades of pink and red. We modified this dataset in two ways to create the MBM dataset (Modified BM). First the $1,200 \times 1,200$ images were cropped to $600 \times 600$ in order to process the images in memory on the GPU and also to smooth out evaluation errors during training for a better comparison. This yields a total of 44 images containing 126 $\pm$ 33 cells (identified nuclei).  In addition, the ground truth annotations were updated after visual inspection to capture a number of unlabeled nuclei with the help of domain experts.

\textbf{Adipocyte Cells}: 
Our final dataset is a human subcutaneous adipose tissue dataset obtained from the Genotype Tissue Expression Consortium (GTEx) \cite{lonsdale2013genotype}. 200 Regions Of Interest (ROI) representing adipocyte cells were sampled from high resolution histology slides by using a sliding window of 1700 $\times$ 1700. Images were then down sampled  to 150 $\times$ 150, representing a suitable scale in which cells could be counted using a 32 $\times$ 32 receptive field. The average cell count across all images is 165$\pm$44.2. Adipocytes can vary in size dramatically (20-200$\mu$) \cite{mclaughlin2014subcutaneous} and given they are densely packed adjoining cells with few gaps, they represent a difficult test-case for automated cell counting procedures.

\begin{table*}[htbp]
\small
\centering
\caption{Comparison of test set mean absolute error (MAE) of counts per image with prior work. Out of all images in each dataset, $N$ images are randomly selected for the training set, $N$ for the validation set, and a fixed size is used for the testing set. At least 10 runs using different random splits and different network initializations are used to calculate the mean and standard deviation.}
\label{tab:overall}

\textbf{VGG Cells} (200 Images Total)
{%

\begin{tabular}{c c c c c}
\toprule
Method & $N=8$ & $N=16$ & $N=32$ & $N=50$\\
\midrule
Predict Average Count & 
$52.5 \pm 2.4 $ & 
$52.5 \pm 2.3$ & 
$52.2 \pm 2.3$ &  
$52.1 \pm 2.4$\\

Cell Profiler & 
\multicolumn{4}{c}{$-- 7.9 \pm 0.3 --$} \\
Lempitsky and Zisserman (2010) & 
$4.9 \pm 0.7 $ & 
$3.8 \pm 0.2$ & 
$3.5 \pm 0.2$ &  
$N/A$\\
Fiaschi et al. (2012) &  
$3.4\pm 0.1$ & 
$N/A$ & 
$3.2 \pm 0.1$ &  
$N/A$\\
Arteta et al. (2014) & 
$4.5\pm 0.6$ & 
$3.8\pm 0.3$ & 
$3.5\pm 0.1$ &  
$N/A$\\
FCRN-A, Xie (2016) & 
$3.9 \pm 0.5$ & 
$3.4 \pm 0.2$ & 
$2.9 \pm 0.2$ & 
$2.9 \pm 0.2${\color{red}\textsuperscript{*}} \\
Count-ception (Proposed) & 
$3.9 \pm 0.4$ & 
$2.9 \pm 0.5$ & 
$2.4 \pm 0.4$ & 
\textbf{$2.3 \pm 0.4$}\\ 
\bottomrule
\multicolumn{5}{p{10cm}}{
\vspace{-5pt}
{\color{red}\textsuperscript{*}}\footnotesize Reported in their work as $N=64$.}
\end{tabular}

}
  
  \vspace{10pt}

  \textbf{MBM Cells} (44 Images Total)
  {%
  
\begin{tabular}{c c c c}
\toprule
Method & $N=5$ & $N=10$ & $N=15$\\
\midrule
Predict Average Count &
$29.4 \pm 2.3$ & 
$28.6 \pm 1.6$ & 
$28.2 \pm 1.6$\\

Cell Profiler -single & \multicolumn{3}{c}{$ -- 19.8 \pm 4.2 --$} \\
Cell Profiler -multiple{\color{red}\textsuperscript{**}} & \multicolumn{3}{c}{$ -- 12.8 \pm 3.1 --$} \\

FCRN-A, Xie (2016) & 
$28.9 \pm 22.6$ & 
$22.2 \pm 11.6$ & 
$21.3\pm 9.4$ \\
Count-ception (Proposed) &
$12.6 \pm 3.0$ & 
$10.7 \pm 2.5$ & 
\textbf{$8.8 \pm 2.3$} \\
\bottomrule
\multicolumn{4}{p{10cm}}{
\vspace{-5pt}
{\color{red}\textsuperscript{**}}\footnotesize Cell Profiler results were obtained using a single pipeline (single) and using three different pipelines (multiple) to account for color differences in two of the eleven images.}
\end{tabular}
}

\vspace{10pt}

\textbf{Adipocyte Cells} (200 Images Total)
{%

\begin{tabular}{c c c c}
\toprule
Method & $N=10$ & $N=25$ & $N=50$\\
\midrule
Predict Average Count &
$33.8 \pm 3.1$ & 
$33.6 \pm 3.0$ & 
$33.5 \pm 2.9$\\

Count-ception (Proposed) &
$25.1 \pm 2.9$ & 
$21.9 \pm 2.8$ & 
$19.4 \pm 2.2$ \\
\bottomrule
\end{tabular}

  }
\end{table*}

\section{Experiments}

First, we compare the overall performance of our proposed model to existing approaches in Table \ref{tab:overall} for each dataset. For each dataset we follow the evaluation protocol used by Lempitsky and Zisserman in 2010 that has been used by all future papers. In this evaluation protocol, training, validation, and testing subsets are used. The held-out testing set size is fixed for all experiments while training and validation sizes ($N$) are varied to simulate lower or higher numbers of labeled examples. The algorithm trains on the training set only while being able to early stop by evaluating its performance on the validation set. The size of the training and validation sets are varied together for simplicity. 

The results of the algorithm using at least 10 random splits are computed and we present the mean and standard deviation. The testing set size remains constant in order to provide constant evaluation. If the testing set were chosen to be all remaining examples ($|$Testing$|$ =  $|$Total$|-2N$) instead of a fixed size then smaller $N$ values would be less impacted by difficult examples in the test set because examples are not sampled with replacement.

As a practitioner baseline comparison we compare our results to Cell Profiler's \cite{Carpenter2006} which uses segmentation to perform object identification and counting. This is representative of how cells are typically counted in biology laboratories.  To do so, we designed two main different pipelines and evaluated the error on 10 splits of 100 randomly chosen images for the synthetic dataset (VGG Cells) and on 10 splits of 10 images for the bone marrow dataset (MBM Cells) to mimic the experimental setup in place since Cell Profiler does not use a training set.  For the MBM Cells, we report the performance using the same pipeline (single) for all images and using three slightly modified versions of the pipeline (multiple) where a parameter was adjusted to account for color differences seen in 8 of the 44 images.

Among other methods we compare with Xie's FCRN-A network \cite{Xie2016}. Only Xie's and our method (Count-ception) are neural network based approaches. Our network is sufficiently deeper than the Xie's FCRN-A network and that representational power together with our redundant counting we are able to perform significantly better. We show in \S \ref{sec:redundant} that the performance of our model matches that of Xie's when the redundant counting is disabled by changing the stride to eliminate redundant counting.

\subsection{Training}

In order to train the network we used the Adam optimization technique \cite{Kingma2014a} with a learning rate of 0.005 and a batch size of 4 images. The training runs for 1000 epochs and the best model based on the validation set error is evaluated on the test set. The weights of the network were initialized using the Glorot initialization method \cite{Glorot2010} adjusted for ReLU gain.

\subsection{Redundant Counting}
\label{sec:redundant}

We claim redundant counting is significant to the success of the method. By increasing the stride we can reduce double counting until there is none. We present the reader Table \ref{tab:stride} which indicates that a stride of 1, meaning the maximum amount of redundant counting patch\_size$^2$, is the optimal choice. As we increase the stride to equal the patch size where no redundant counting is occurring the accuracy is reduced.

The power of this algorithm is in the redundant counting. However, increasing the redundant count is complicated. The receptive field could be increased but this will add more parameters which cause the network to overfit the training data. We explored a receptive field of 64x64 and found that it did not perform better. Another approach could be to use dilated convolutions \cite{Yu2016} which would be equivalent to scaling up the input image resolution.

\begin{table}[]
\begin{center}
\caption{Comparison of different strides ($s$) in order to reduce the redundant counting. Results are compared using the mean absolute error of the output predictions. For these experiments we use $N=32$ examples. Here Train \& Test means the stride was set at that value for training and testing. Having a larger stride during training means seeing less data. A network trained with $s=32$ has seen 32 times less data that with $s=1$. In the Test Only case the network was trained with $s=1$ and then evaluation on the test set was limited to different strides so less redundant predictions are made.}
\begin{tabular}{c c c c c}
\toprule
Stride & $s=1$ & $s=8$ & $s=16$ & $s=32$ \\
\midrule
\hspace{-5pt}Train \& Test & 
2.4$\pm$0.4 & 
3.5$\pm$0.1 & 
4.0$\pm$0.2 & 
5.2$\pm$0.4 \\
Test Only & 
2.4$\pm$0.4 & 
2.5$\pm$0.4 & 
2.7$\pm$0.3 & 
3.0$\pm$0.3 \\
\bottomrule
\end{tabular}

\label{tab:stride}

\end{center}
\end{table}

\subsection{Runtime and Implementation}

The run-time of this algorithm is not trivial. We explored models with less parameters and found they could not achieve the same performance. Shorter models (fewer layers) or narrower models (less filters per layer) tended to not have enough representational power to count correctly. Making the network wider would cause the model to overfit. The complexity of the Inception modules were significant to the performance of the model.

The network was implemented in lasagne (version 0.2.dev1) \cite{lasagne} and Theano (version 0.9.0rc2.dev) \cite{theano} using the libgpuarray backend. The source code and data will be made available online\footnote{\url{https://github.com/ieee8023/countception}}.

\section{Conclusion}

In this work we rethink the density map method by Lempitsky and Zisserman \cite{Lempitsky2010} and instead predict counts in a redundant fashion in order to average over errors and reduce overfitting. This \textit{redundant counting} approach merges ideas by Segui \cite{Segui2015} of networks that count everything in their receptive field with ideas by Lempitsky and Zisserman of using the density map of objects together with ideas by Xie \cite{Xie2016} and Arteta \cite{Arteta16} of using fully convolutional processing.

We call our new approach Count-ception because our approach utilizes a counting network internally to perform the \textit{redundant counting}. We demonstrate that this approach outperforms existing approaches and can also perform well with very complicated cell structure even where the cell walls adjoin other cells. This approach is promising for tasks with different sizes of objects which have complicated structure. However, the method has some limitations. Although the count map can be used for localization it cannot easily provide $x,y$ locations of objects.

\section{Acknowledgments}
This work is partially funded by a grant from the U.S. National Science Foundation Graduate Research Fellowship Program (grant number: DGE-1356104) and the Institut de valorisation des donn\'{e}es (IVADO). This work utilized the supercomputing facilities managed by the Montreal Institute for Learning Algorithms, NSERC, Compute Canada, and Calcul Queb\'{e}c. We also thank NVIDIA for donating a DGX-1 computer used in this work. 

{\small
\bibliographystyle{ieee}
\bibliography{cohen,neuralnetworks,computervision}
}

\end{document}